\documentclass[10pt,twocolumn,letterpaper]{article}

\usepackage{cvpr}              % To produce the CAMERA-READY version
\usepackage{graphicx}
\usepackage{amsmath}
\usepackage{amssymb}
\usepackage{booktabs}
\usepackage[ruled,linesnumbered]{algorithm2e}

\usepackage[pagebackref,breaklinks,colorlinks]{hyperref}

\usepackage[capitalize]{cleveref}
\crefname{section}{Sec.}{Secs.}
\Crefname{section}{Section}{Sections}
\Crefname{table}{Table}{Tables}
\crefname{table}{Tab.}{Tabs.}

\def\cvprPaperID{114514} % *** Enter the CVPR Paper ID here
\def\confName{CVPR}
\def\confYear{2022}

\begin{document}

\title{BERT-Enhanced Retrieval Tool for Homework Plagiarism Detection System}

\author{Jiarong Xian*\\
South China Normal University \\
{\tt\small 20214001019@m.scnu.edu.cn}
\and
Jibao Yuan*\\
South China Normal University\\
{\tt\small 20214001082@m.scnu.edu.cn}
\and
Peiwei Zheng*\\
South China Normal University\\
{\tt\small 20212005173@m.scnu.edu.cn}
\and
Dexian Chen*\\
South China Normal University\\
{\tt\small 20202133022@m.scnu.edu.cn}
\and
Yuntao Nie*\\
South China Normal University\\
{\tt\small 20212005369@m.scnu.edu.cn}
}

\maketitle

%%%%%%%%% ABSTRACT
\begin{abstract}

    Text plagiarism detection task is a common natural language processing task that aims to detect whether a given text contains plagiarism or copying from other texts. In existing research, detection of high level plagiarism is still a challenge due to the lack of high quality datasets. In this paper, we propose a plagiarized text data generation method based on GPT-3.5, which produces \textbf{32,927} pairs of text plagiarism detection datasets covering a wide range of plagiarism methods, bridging the gap in this part of research. Meanwhile, we propose a plagiarism identification method based on Faiss with BERT with high efficiency and high accuracy. Our experiments show that the performance of this model outperforms other models in several metrics, including \textbf{98.86\%}, \textbf{98.90\%}, \textbf{98.86\%}, and \textbf{0.9888} for Accuracy, Precision, Recall, and F1 Score, respectively. At the end, we also provide a user-friendly demo platform that allows users to upload a text library and intuitively participate in the plagiarism analysis.
\end{abstract}

%%%%%%%%% BODY TEXT
\section{Introduction}
    You are a teacher who is responsible for detecting plagiarism in assignments submitted by students. Plagiarism in assignments can be categorised into three types\cite{clough2011developing}:
    \begin{enumerate}
        \setlength{\itemsep}{0pt}
        \setlength{\parskip}{0pt}
        \item \textbf{word-by-word plagiarism}: It's the equivalent of copying;
        \item \textbf{Imitation plagiarism}: Plagiarizing by rewriting words and grammar;
        \item \textbf{Creative plagiarism}: Plagiarism is achieved by reusing and rewriting original ideas from the source text;
    \end{enumerate}

    Obviously, for simple plagiarism like word-by-word plagiarism, we can detect it by some simple methods such as keyword search. 
    For Imitation plagiarism and Creative plagiarism, two plagiarisms that are more difficult to recognize, fine-tuned Large Language Models (LLMs) are generally used for detection.

    LLMs process a large amount of textual data in the pre-training phase, including a variety of literary works, academic papers, and web content. This enables them to learn and understand a variety of linguistic expressions, including different writing styles and language structures. As a result, they can recognize texts that have undergone significant changes at the lexical and grammatical levels. For example, during word embedding, the BERT model takes into account the context to the left and right of a given word in a sentence. This bi-directional contextual understanding allows the model to better capture the deeper semantics of the sentence.

    Reasoning with LLMs consumes a lot of computational resources. Therefore, we need a fast retrieval method to filter out superficially similar texts that may involve plagiarism. FAISS\cite{douze2024faiss} is a good choice for storing and retrieving embedding vectors for BERT.

    Then, we can splice the retrieved results and let it go through an MLP to derive the probability of plagiarism occurrence.

    However, the fact is that it is difficult to obtain a real dataset of plagiarized assignments. The reason is simple: no child who plagiarizes homework wants to be caught by the teacher.

    Fortunately, the development of large-scale language models (LLMs) such as GPT-3.5 has opened up new avenues for natural language processing, especially in terms of data augmentation for low-resource classification tasks. So, we used Prompt to let GPT-3.5 plagiarize some AI papers on arxiv, and came up with a plagiarism compartmentalized diverse dataset.

    Finally, we trained a model for identifying plagiarism categories and it performed far better than other models on this dataset.

    These are some of our main contributions:
    \begin{enumerate}
        \setlength{\itemsep}{0pt}
        \setlength{\parskip}{0pt}
        \item \textbf{Dataset}: We used GPT-3.5 to generate a plagiarized corpus with diverse plagiarism styles;
        \item \textbf{Accurate and efficient plagiarism identification}: We propose a plagiarism identification framework to quickly and accurately retrieve and determine the type of plagiarism;
        \item \textbf{Our demo}: We built a simple demo where users are able to upload text libraries and perform plagiarism analysis in a visual way;
    \end{enumerate}
    
\section{Related works}
\subsection{Detection Tool}
In the context of plagiarism detection using language models, we consider the BERT model as a black box. For initial screening, FAISS is used to identify potential plagiarized texts. These candidate texts are then input into the BERT model for a more in-depth semantic analysis to determine if plagiarism exists between two target texts.

FAISS provides crucial support in the preliminary screening of assignment submissions, employing efficient similarity search algorithms. Its GPU-accelerated searches optimize k-nearest neighbors searches based on L2 distance, essential for handling large-scale datasets.\cite{douze2024faiss} FAISS's product quantization code enhances this process, offering a more effective and memory-efficient method than binary codes. Its IVFADC index structure is particularly adept at managing large datasets, aligning well with the preliminary screening requirements in plagiarism detection tasks.

Sentence BERT, especially in its MPNet-based improved version\cite{song2020mpnet}, marks a significant advancement in natural language processing. It offers profound semantic understanding of text representation, which is key for identifying not only direct plagiarism but also complex rewrites and subtle textual modifications. Initially used for sentence-level comparisons, SentenceBERT's capability to recognize deep semantic relationships within texts is crucial for the detailed analysis of cases flagged by FAISS. In its updated version, SentenceBERT also excels in generating semantic vectors for longer paragraph-level texts. In our research, SentenceBERT is used to generate semantic segments of FAISS assignment vectors during the preliminary screening phase. The BERT model is then fine-tuned on our dataset, enhancing its effectiveness in downstream tasks like plagiarism detection.

\subsection{Dataset Construction}
The majority of datasets constructed for plagiarism detection research employ a binary labeling approach for samples, categorizing them as either plagiarized or not, as discussed by Alvi et al., (2021)\cite{alvi2021paraphrase}. 

However, varying degrees of plagiarism are evident in different samples, depending on the methods used (such as synonym replacement, grammatical structure alteration, semantic sequence modification) and the frequency of these modifications. Clough et al., (2011)\cite{clough2011developing} created a plagiarism corpus reflecting different degrees of plagiarism, offering a model for how to categorize plagiarized samples in dataset construction. Their dataset, with approximately 500 samples, is available for download here (\url{	https://github.com/josecruzado21/plagiarism_detection}). 

Mehrafarin et al.,(2022)\cite{mehrafarin2022importance} examined the required dataset size for fine-tuning the BERT model, finding in experiments across five different classification tasks that beyond a dataset size of 7k, the cost-benefit ratio of enhancing model performance by increasing dataset size diminishes. This suggests that an appropriately sized dataset should be several thousand samples or more. Therefore, a larger classified dataset is needed. Due to ethical and legal considerations, methods for obtaining plagiarism corpus datasets are almost exclusively limited to machine construction and manual simulation, as noted by Alvi et al., (2021)\cite{alvi2021paraphrase}. 

The work of Møller et al., (2023)\cite{moller2023prompt} demonstrates the feasibility of using GPT-4 for dataset construction, highlighting that for more complex tasks, designing intricate prompts allows GPT-4 constructed datasets to rival those created by humans. The study by Bsharat et al., (2023)\cite{bsharat2023principled} in December 2023 summarized 26 principles for writing GPT prompts, while Zhou et al., (2022) updated and summarized comprehensive text paraphrasing studies related to plagiarism, including 24 specific paraphrasing techniques. This provides detailed guidance for instructing GPT on how to paraphrase.

\section{Data description}
    Based on Professor Li Mu's recommendation, we downloaded PDF files of 78 AI papers from ARXIV. Through manual screening, we extracted an average of 60 text fragments for each paper (excluding parts such as figures, related work and appendices). Considering the 512-tokens limit of SBERT, we segmented the excessively long segments, and finally nearly 5,000 segments of text were used as the raw data for this work. (\url{https://arxiv.org/})

\section{Method}
    
\subsection{GPT-3.5-based data enhancement}
    Here, we have used five methods to enhance the data. Among them are negative sampling, chaotic rearrangement and three Prompt-based GPT-3.5 enhancement schemes.

    The data were categorised into five types according to different methods. Specifically, these types include "plagiarism", which involves only a change in sentence order; "minor revision", which involves synonym substitution; "moderate revision", which combines synonym substitution and a change in grammatical structure; "viewpoint plagiarism", which involves synonym substitution, a change in the order of grammar and information, and stylistic modifications; and "no plagiarism", which involves texts from different authors on the same topic. The 'plagiarism' and 'no plagiarism' levels were compiled using an automated method, while the 'mild', 'moderate' and 'opinion' plagiarism categories were compiled using the GPT-3.5 API.

    Eventually, we arrive at 32,927 text pairs of the form $(t_1,t_2,label)$. Where $t_1$ is the original text, $t_2$ is the text to be detected, and $label$ denotes the type of plagiarism of $t_2$ on $t_1$.

    Here, $label$ has three values, which are:
    \begin{enumerate}
        \setlength{\itemsep}{0pt}
        \setlength{\parskip}{0pt}
        \item \textbf{No Plagiarism}: Generated by negative sampling;
        \item \textbf{Imitation Plagiarism}: Generated through GPT-3.5;
        \item \textbf{Shuffle Plagiarism}: Generated through random disruption;
    \end{enumerate}

    A more detailed Prompt methodology can be found in the Appendix.

\subsection{Text plagiarism identification model}
    Our model is divided into two parts, the feature representation module and the classifier. See Figure 1 for details.

    \begin{figure}[h]
        \centering
        \includegraphics[width=0.75\linewidth]{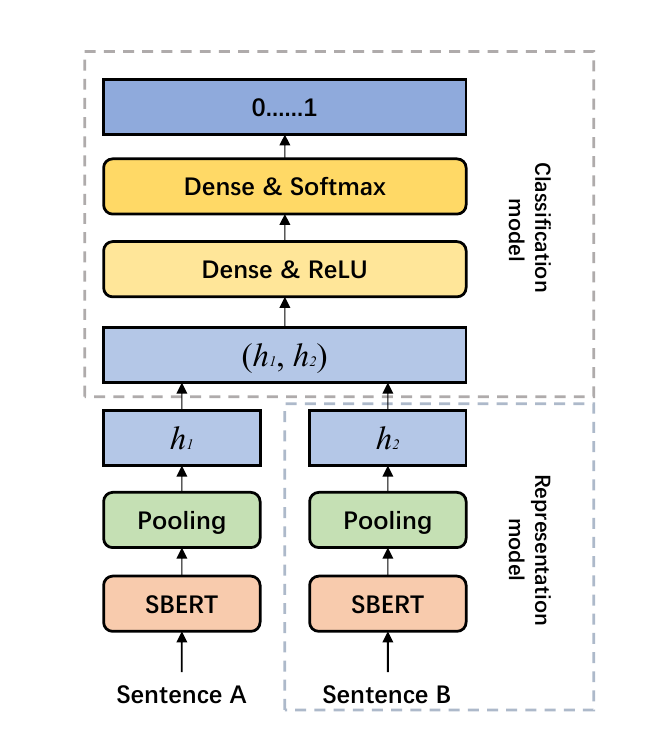}
        \caption{Our model chart}
        \label{fig:enter-label}
    \end{figure}

\subsubsection{SBERT-based text feature representation}
    \hspace{0.4cm} 
    SBERT (Sentence-BERT) is an approach for textual feature representation based on an improved version of the BERT (Bidirectional Encoder Representations from Transformers) model specifically designed for generating embedding vectors at the sentence level.
    
    The advantage of SBERT is that by encoding text as high-quality embedding vectors, it can provide better performance in a variety of natural language processing tasks.

    Here, SBERT will act as a text feature extractor (Representation Model) to transform the input text $t$ into a high latitude vector $h \in R^{n}$ for subsequent plagiarism identification tasks. See equation (1)(2)(3) for details.
    \begin{equation}
        \tilde{t}=tokenizer\left( t \right) \\
        \label{eq:representation-model-tokenizer}
    \end{equation}
    \begin{equation}
        e=embedding\left( \tilde{t} \right) 
        \label{eq:representation-model-embedding}
    \end{equation}
    \begin{equation}
        h=Pool_{[CLS]}\left( SBERT\left( e \right) \right) 
        \label{eq:representation-model-pool}
    \end{equation}
    
    where $t$ is the input text, $\tilde{t}$ is the result after tokeniser, $e$ is the word embedding result and $h$ is the [CLS] value of SBERT.

    For ease of subsequent representation and understanding, the feature representation model is denoted as  $model_r$, where the input is $t$ and the output is $h$. Where $t$ is the text and $h$ is the feature vector of $t$.
    
\subsubsection{MLP-based text plagiarism classifier}
    \hspace{0.4cm}
    Given $(t_1,t_2)$ , where $t_1$ is the original text and $t_2$ is the text to be detected, our task is to determine whether $t_2$ plagiarised $t_1$. First, the two texts are represented as feature vectors , denoted as $(h_1,h_2)$. For details, see Eq. (4).
    \begin{equation}
        \left( h_1,h_2 \right) =model_r\left( \left( t_1,t_2 \right) \right) 
        \label{eq:classification-model-represent}
    \end{equation}
    
    Next, denote $concat(h_1, h_2)$ as $u$. $u$ through a dense layer and ReLU layer, denoted as $\tilde{u}$. For details, see Eq. (5).
    \begin{equation}
        \tilde{u}=\mathrm{ReLU}\left( uW_1+b_1 \right) 
        \label{eq:classification-model-mlp-1}
    \end{equation}

    Finally, $\tilde{u}$ through a dense layer and Softmax layer, denoted as $p$. where $p_i$ denotes the probability that the $i_{th}$ plagiarism occurs. For details, see Eq. (6).
    \begin{equation}
        p=\mathrm{Softmax} \left( \tilde{u}W_2+b_2 \right) 
        \label{eq:classification-model-mlp-2}
    \end{equation}

    where $W_1$, $W_2$, $b_1$, $b_2$ is a learnable parameter, Softmax is detailed in Eq. (7), and ReLU is detailed in Eq. (8).
    \begin{equation}
        \mathrm{Softmax} \left( z_i \right) =\frac{\exp \left( z_i \right)}{\sum{z_j}}
        \label{eq:Softmax}
    \end{equation}
    \hspace{0.4cm} where $z_i$ is the $i_{th}$ element of the input vector $z$.
    \begin{equation}
        \mathrm{ReLU} \left( x \right) = max(x,0)
        \label{eq:ReLU}
    \end{equation}

    For ease of subsequent representation and understanding, the feature classification model is denoted as  $model_c$, where the input is $(h_1, h_2)$ and the output is $p \in R^3$. where $h_1$ and $h_2$ are the feature vectors of the two texts, respectively, and $p_i$ is the probability of occurrence of type $i$ plagiarism. 
    
\subsubsection{Our training algorithm}
    \hspace{0.4cm}
    Cross-Entropy loss function is a metric used to measure the difference between two probability distributions. In deep learning, Cross-Entropy is commonly used to measure the difference between the model output and the true label, and is one of the common loss functions in classification tasks. We choose the cross-entropy loss function as the objective function for model training. For details, see Eq. (9).
    \begin{equation}
        CE\left( p,q \right) =-\sum_{i=1}^C{q_i\log \left( p_i \right)}
        \label{eq:cross-entry-loss}
    \end{equation}
    
    where $q_i$ is the true label and $p_i$ is the predicted probability.

    Our training algorithm is detailed in Algorithm 1.

    \begin{algorithm}
        \SetAlgoLined
        \caption{Training algorithm}
        \KwData{Dataset $\mathcal{D} $, such as $(t_1, t_2, label)$; \\ \hspace{1cm}Model parameters $\theta$;}
        \KwResult{Parameters after training $\theta$; }
        \For{$(t_1, t_2, labels) \in \mathcal{D}$}{
            calculate $(h_1, h_2) \longleftarrow model_r((t_1, t_2))$ \;
            calculate $logits \longleftarrow model_c(h_1, h_2)$ \;
            get loss $L \longleftarrow CE(logits, labels)$ \;
            update parameters $\theta \longleftarrow backward(L, \theta)$ \;
        }
    \end{algorithm}

\subsection{Faiss-based text retrieval}
    Given the text to be detected, denoted as $t$. It is difficult to traverse a large number of texts for plagiarism detection.
    Therefore, we need a way to quickly retrieve some texts with a high likelihood of plagiarism. Then, we can use the plagiarism detection model to make a judgement.

    FAISS (Facebook AI Similarity Search) is an open source library developed by Facebook for efficient similarity search. It is mainly used to process large-scale high-dimensional vectors, such as images, text and other embedded representations. At the same time, FAISS supports storage, retrieval and other functions. We can use the faiss framework to accelerate text retrieval and reduce the scope of plagiarism detection analysis.

    For a retrieval task, Faiss can be divided into the following three steps: creating index table, clustering training, querying. 
    
    The core of faiss is the index, which helps to retrieve vectors efficiently. The vector dimension we choose is the output dimension \textbf{768} of SBERT, and the quantizer is the vector \textbf{dot product}.

    We cannot accept the huge time consuming nature of traversal retrieval. Therefore, we use Faiss's \textbf{inverted file} to speed up the search. Its idea is to use \textbf{k-means} algorithm to establish the clustering centre in advance, so that the search can query the nearest centre, and then compare all the vectors in the clusters to get the similar vectors, which reduces the search scope. 

    In addition, all the vector data of the index table is stored in memory, in order to handle more text data with smaller memory footprint, faiss provides a compression algorithm based on Product Quantizer.

    Its algorithm steps are as follows:

    \begin{enumerate}
        \setlength{\itemsep}{0pt}
        \setlength{\parskip}{0pt}
        \item Cut a feature vector into multiple segments;
        \item Perform k-means clustering in the subspace represented by each segment;
        \item Use the number of the closest clustering centre in each segment as the index value, i.e., the feature vector is compressed into a new vector consisting of the number of the clustering centre in each sub-segment;
        \item Memory retains only the compressed vectors and cluster centres.
    \end{enumerate}
    
    This reduces the memory footprint and speeds up the distance calculation. The whole process above can be described as Algorithm 2. 

    \begin{algorithm}
        \SetAlgoLined
        \caption{Faiss-based texts retrieve algorithm}
        \KwData{Paper segments from arvix $\mathcal{S}$; \\ \hspace{1cm}Representation model $model_r$;}
        \KwResult{Faiss's index $I$}
        Get all embeddings $E \longleftarrow model_r(\mathcal{S})$ \;
        Create Faiss Index $I$ \;
        Train Index $I$ with embeddings $E$ (Product Quantizer,k-means) \;
        Add all embeddings $E$ to Index $I$ \;
    \end{algorithm}

    After processing, we obtain a Product Quantizer optimized and k-means trained Faiss index table $I$. For an input vector $h$ , Faiss is able to retrieve the $top-k$ alternative vectors that are most similar to $h$ , denoted as $[v_1, v_2, \dots , v_k]$.

\section{Experiments}
\subsection{Metrics}
Some common metrics for classification task evaluation are used here, which are usually computed using four values in the confusion matrix: true positive (TP), true negative (TN), false positive (FP), and false negative (FN).

$Accuracy$ is the number of samples correctly predicted by a classification model as a proportion of the total number of samples. It is an intuitive metric, but may not be a good performance measure in the case of unbalanced categories, as the model may tend to predict the majority of categories and ignore the minority. For details, see Eq. (10).

\begin{equation}
    Accuracy=\frac{TP+TN}{TP+TN+FP+FN}
    \label{eq:accuracy}
\end{equation}

$Precision$ is the proportion of samples predicted by the model to be positive cases that are actually positive cases. It measures how many of the model's predictions are correct and is an important metric for tasks where false positives are not tolerated. For details, see Eq. (11).

\begin{equation}
    Precision=\frac{TP}{TP+FP}
    \label{eq:precision}
\end{equation}

$Recall$ is the proportion of samples that are actually positive examples that the model successfully predicts as positive. Recall is a key metric in tasks where the omission of positive examples cannot be tolerated. For details, see Eq. (12).

\begin{equation}
    Recall=\frac{TP}{TP+FN}
    \label{eq:recall}
\end{equation}

$F1$ score is the reconciled mean of precision and recall, which provides a comprehensive consideration of the model's performance.The F1 score is particularly useful when dealing with data from unbalanced categories, as it takes into account both false positives and omissions from the model. For details, see Eq. (13).

\begin{equation}
    F1=2\times \frac{Precision\times Recall}{Precision+Recall}
    \label{eq:f1}
\end{equation}

\subsection{Training details}
In this section, we'll tell you more details about training.

\textbf{Training-testing set division.} We divide the dataset $\mathcal{D}$ into $\mathcal{D}_{train}$ and $\mathcal{D}_{test}$ in the ratio of 8:2, where $\mathcal{D}_{train}$ is the training set and $\mathcal{D}_{test}$ is the test set.

\textbf{Model parameter setting.} In this experiment, we tried to train models with 4 different structures. For details, see table 1.
\begin{table}[h]
    \caption{Model structure details}
    \centering
    \begin{tabular}{cccc}
        \toprule
        Model & BERT & pool & hidden size\\
        \midrule
        Ours-SBERT    & sbert-based  & [CLS] &  $\left[ \begin{array}{c}	1536\times 512\\	512\times 3\\\end{array} \right] $ \\
        Ours-BERT     & bert-based   & [CLS] &  $\left[ \begin{array}{c}	1536\times 512\\	512\times 3\\\end{array} \right] $  \\
        SBERT-COS     & sbert-based  & [CLS] & -  \\
        SBERT-DOT     & sbert-based  & [CLS] & - \\
        \bottomrule
    \end{tabular}
    \label{tab:three-line-table}
\end{table}

\textbf{Optimizer setting.} We used the \textbf{Adam} optimizer for model training, where the \textit{learning rate} was set to 0.001, the \textit{batch size} was set to 1024 and the \textit{maximum number of iteration} rounds was set to 20. 

In the end, Ours-SBERT reached convergence after 6 hours of training and Ours-BERT reached convergence after 5 hours of training on a single NVIDIA GeForce RTX 3070. 

\subsection{Evaluation results}

\subsubsection{Fine-tuning Model Evaluation}
    \hspace{0.4cm}
    We evaluated four models. For details, see table 2. The performance of our proposed Ours-SBERT model is outstanding and outperforms other models. 

    \begin{table}[h]
        \caption{Fine-tuning Model Evaluation results}
        \centering
        \begin{tabular}{ccccc}
            \toprule
            Model & Accuracy & Precision & Recall & F1 Score \\
            \midrule
            Ours-SBERT    & \textbf{98.86\%} & \textbf{98.90\%} & \textbf{98.86\%} & \textbf{0.9888} \\
            Ours-BERT     & 95.31\% & 95.59\% & 95.31\% & 0.9545 \\
            SBERT-COS     & 78.14\% & 78.16\% & 78.14\% & 0.7815 \\
            SBERT-DOT     & 80.04\% & 78.84\% & 80.04\% & 0.7944 \\
            \bottomrule
        \end{tabular}
        
        \label{tab:three-line-table}
    \end{table}

\subsubsection{Faiss Performance Evaluation}
    \hspace{0.4cm}
    Since the use of Product Quantizer(PQ) optimisation and k-means training acceleration affects the matching accuracy, we need to examine the index table and compare the different optimisation methods. 

    So we train Index with different optimizations which are:
    \begin{enumerate}
        \setlength{\itemsep}{0pt}
        \setlength{\parskip}{0pt}
        \item no clustering and memory optimization;
        \item k-means clustering acceleration (100 clustering centers, alternative 20 clustering centers);
        \item k-means clustering acceleration and PQ memory optimization (100 clustering centers, alternative 20 clustering centers, each segmentation vector dimension is 16);
    \end{enumerate}

    We train Index $I$ with different optimisations and use the shuffle plagiarism processed text for match checking, if the returned $top-10$ segments contain the original text, the match is considered successful, otherwise the match is considered failed. Then we calculate the matching success rate, see Table 3 for details.
    
    \begin{table}[h]
        \caption{Faiss Evaluation results}
        \centering
        \begin{tabular}{cccc}
            \toprule
            Strategy & Success Rate & Time(ms/vector) & Dim \\
            \midrule
            Nothing    & \textbf{99.69\%} & 0.1826 & 768\\
            K-Means     & 99.60\% & \textbf{0.0651} & 768\\
            K-Means\&PQ     & 99.53\% & 0.2938 & \textbf{16} \\
            \bottomrule
        \end{tabular}
        \label{tab:three-line-table}
    \end{table}

    The results show that the index $I$ accelerated by k-means training takes less time while guaranteeing the success rate of matching. It can meet our needs of text retrieval and reducing the scope of plagiarism detection analysis. When memory is scarce, Product Quantizer can also be used to optimize memory.
    
\subsection{Visualisation}
    The following is a sample example, including inputs, intermediate results, and outputs.
    
    \textbf{Input}: For these trials, we utilize the identical network framework as mentioned in [22]; that is, two VGG-M-2048 networks [3]. The combination layer is inserted at the final convolutional layer, following rectification, meaning its input comes from the result of ReLU5 from the two channels. This choice was made based on superior outcomes in initial trials compared to options such as the non-rectified output of conv5. At this juncture, the attributes are already significantly informative while also offering rough location details. Following the combination layer, a solitary processing channel is employed.

    \textbf{top-1}: To be clear, our intention here is to fuse the two networks (at a particular convolutional layer) such that channel responses at the same pixel position are put in correspondence. To motivate this, consider for example discriminating between the actions of brushing teeth and brushing hair. If a hand moves periodically at some spatial location then the temporal network can recognize that motion, and the spatial network can recognize the location (teeth or hair) and their combination then discriminates the action.

    \textbf{top-2}: For these experiments we use the same network architecture as in [22]; i.e. two VGG-M-2048 nets [3]. The fusion layer is injected at the last convolutional layer, after rectification, i.e. its input is the output of ReLU5 from the two streams. This is chosen because, in preliminary experiments, it provided better results than alternatives such as the non-rectified output of conv5. At that point the features are already highly informative while still providing coarse location information. After the fusion layer a single processing stream is used.

    \textbf{top-3}: Our proposed architecture (shown in Fig. 4) can be viewed as an extension of the architecture in Fig. 2 (left) over time. We fuse the two networks, at the last convolutional layer (after ReLU) into the spatial stream to convert it into a spatiotemporal stream by using 3D Conv fusion followed by 3D pooling (see Fig. 4, left). Moreover, we do not truncate the temporal stream and also perform 3D Pooling in the temporal network (see Fig. 4, right). The losses of both streams are used for training and during testing we average the predictions of the two streams. In our empirical evaluation (Sec. 4.6) we show that keeping both streams performs slightly better than truncating the temporal stream after fusion.

    \textbf{top-4}: In this section we report results using a 34-layer version of R(2+1)D, which we denote as R(2+1)D-34. The architecture is the same as that shown in the right column of Table 1, but with 3D convolutions decomposed spatiotemporally in (2+1)D. We train our R(2+1)D architecture on both RGB and optical flow inputs and fuse the prediction scores by averaging, as proposed in the two-stream framework [29] and subsequent work [4], [9]. We use Farneback's method [8] to compute optical flow because of its efficiency.

    \textbf{top-5}: This spatial correspondence is easily achieved when the two networks have the same spatial resolution at the layers to be fused, simply by overlaying (stacking) layers from one network on the other (we make this precise below). However, there is also the issue of which channel (or channels) in one network corresponds to the channel (or channels) of the other network.

    The detailed judgment results are shown in Table 4.

    \begin{table}[h]
        \caption{Judgment results}
        \centering
        \begin{tabular}{ccccc}
            \toprule
            top-k & class \\
            \midrule
            top-1 & No Plagiarism \\
            top-2 & Imitation Plagiarism \\
            top-3 & No Plagiarism \\
            top-4 & No Plagiarism \\
            top-5 & No Plagiarism \\
            \bottomrule
        \end{tabular}
        \label{tab:three-line-table}
    \end{table}

\section{Discussions}
    Initially, we expected SBERT to be able to distinguish three different degrees of plagiarism (synonym substitution, synonym substitution + grammatical structure modification, synonym substitution + grammatical structure modification + semantic order modification + linguistic style modification). However, the experimental results show that the SBERT model cannot distinguish these three kinds of plagiarism on our dataset. 

    After experimental analyses, we found that the semantic vectors of these three plagiarisms and their similarities make it difficult for the model to distinguish between them.

    In a study by \cite{nikolaev2023representation}, we found out the reason. On the semantic search, SBERT extracts semantic kernel features that are synonymous with a series of nominal elements in the main clause, ignoring words that only serve a grammatical purpose. That is, the model focuses on what things the sentence describes rather than the predicate/comment of the sentence, i.e., what the sentence actually says about those things. Our first experimental attempts corroborate this idea. SBERT still has significant syntactic analytical power, though, leading the SBERT model for semantic search to view noun element synonymy as a semantic feature of the sentence mainly because of the nature of training these SBERT datasets. 

\section{Conclusions}
We designed and implemented the entire system for plagiarism detection and provided visualization demos. the system consists of sentence bert for generating text vectors, Faiss framework for fast matching of vector libraries, fine-tuned bert model for analyzing plagiarism classification, and visualization demos.

We compare the classification accuracy of different bert models, and our evaluation shows that our model is better suited to the task of our dataset and more suitable for plagiarism classification evaluation. We also compared different acceleration schemes and effects of Faiss, and selected the optimal scheme and parameters to apply to our plagiarism detection system.

\section{Team Contribution and Acknowledgement}
\subsection{Team Contribution}
    Our team division of labour and contribution is follow: 
    \begin{enumerate}
        \item \textbf{Jibao Yuan}: Data enhancement, model construction and training, paper writing, 30\%;
        \item \textbf{Jiarong Xian}: Data collection, model construction and training, paper writing, 30\%;
        \item \textbf{Peiwei Zheng}: Project management, system framework, prompt design, data collection, paper writing, 25\%;
        \item \textbf{Dexian Chen}: Data collection, data processing and testing, paper writing, 15\%;
    \end{enumerate}

\subsection{Acknowledgement}
    Thanks to Mu Li for providing the list of papers, and to each team member who collected the data, spent a relatively large amount of time processing the PDF text, segmenting it, and using the GPT API for data enhancement. Thanks to Autodl platform for providing arithmetic resources, and thanks to the instructor for his patience.

%%%%%%%%% REFERENCES
{\small
\bibliographystyle{ieee_fullname}
% \bibliography{ref.bbl}

}

\newpage % 强制分页
\onecolumn % 切换到单栏
\section{Appendix}
\subsection{GPT-3.5 Prompt}

\subsubsection{slight prompt}

\textbf{instruction}: I am building a dataset for plagiarism assignments and need your assistance. Please pretend to be a student who intends to plagiarize, and I will provide you with a plagiarism strategy and source text. Based on my plagiarism strategy, please modify the source text. Then output the modified text to me. Remember, to save your output time, you only need to output the modified text to me. Use $<$start$>$ to indicate the start of the output, and $<$end$>$ to indicate the end of the output.

\textbf{plagiarism strategy}: Replace various words in the source text with synonyms while maintaining the original sentence structure and information sequence.

\textbf{example}: 

Their bank account balance \textbf{reached the maximum insured amount} $\iff$ Their bank account balance was \textbf{at least 250 thousand dollars}

\textbf{Mr. Smith} just bought a new computer $\iff$ \textbf{Bob} just bought a new computer

The program \textbf{runs fast} $\iff$ The program \textbf{does not run slowly}

A spike in sales performance will \textbf{save the company from bankruptcy} $\iff$ Only a spike in sales performance \textbf{will halt the company’s bankruptcy}.

I \textbf{bought} a plane ticket online $\iff$ A plane ticket was \textbf{sold} to me online

Increase in \textbf{salaries} is often a great indicator of performance $\iff$ Increase in \textbf{salary} is often a great indicator of performance.

Is \textbf{this} your own work? $\iff$ Is \textbf{that} your own work?

There are many accounts of that hero’s legacy all \textbf{differing} in perspectives. $\iff$ There are \textbf{different} versions of that hero’s legacy.

These numbers, \textbf{interestingly}, seem to appear in the world around us. $\iff$ These numbers \textbf{interestingly} seem to appear in the world around us.

\textbf{source text}: Input your source text

\subsubsection{obvious prompt}

\textbf{instruction}: I am building a dataset for plagiarism in assignments, and now I need your help. Please pretend to be a student who intends to plagiarize an assignment. I will provide you with plagiarism strategies and source texts to copy from. You should modify the source text according to my plagiarism strategies, and then output the modified text to me.Remember, to save your output time, you only need to output the modified text to me.Use <start> to indicate the start of the output, and <end> to indicate the end of the output.

\textbf{plagiarism strategy}: **Use synonyms, change grammatical structures, and you may alter punctuation, but fundamentally maintain the order of information in the sentences.**

\textbf{example}: Synonym replacement example: 
Their bank account balance \textbf{reached the maximum insured amount} $\iff$ Their bank account balance was \textbf{at least 250 thousand dollars}

\textbf{Mr. Smith} just bought a new computer $\iff$ \textbf{Bob} just bought a new computer

The program \textbf{runs fast} $\iff$ The program \textbf{does not run slowly}

A spike in sales performance will \textbf{save the company from bankruptcy} $\iff$ Only a spike in sales performance \textbf{will halt the company’s bankruptcy}.

I \textbf{bought} a plane ticket online $\iff$ A plane ticket was \textbf{sold} to me online

Increase in \textbf{salaries} is often a great indicator of performance $\iff$ Increase in \textbf{salary} is often a great indicator of performance.

Is \textbf{this} your own work? $\iff$ Is \textbf{that} your own work?

There are many accounts of that hero’s legacy all \textbf{differing} in perspectives. $\iff$ There are \textbf{different} versions of that hero’s legacy.

\textbf{Grammatical structure modification example:}

Does working in that technology company pay well? Does it provide great 401k plans for its employees? $\iff$ They will work at the company to get high pay or to obtain great 401k plans.

How he would stare! $\iff$ He would surely stare!

Jacob \textbf{programmed} the game $\iff$ the game’s \textbf{programmer} is Jacob

“You must finish the project by the end of today,” \textbf{demanded my manager} $\iff$ \textbf{My manager demanded that} I must finish the project by the end of today

Alice presented \textbf{the gift to Bob}. $\iff$ Alice presented \textbf{Bob with the gift}.

All \textbf{spoken languages} are natural languages $\iff$ The \textbf{English language} is a natural language

\textbf{Initially,} we begin with the scientific method $\iff$ we begin with the scientific method \textbf{initially}

\textbf{source text}: Input your source text

\subsubsection{significant prompt}

\textbf{instruction}: I am building a dataset for plagiarism in assignments, and now I need your help. Please pretend to be a student who intends to plagiarize an assignment. I will provide you with plagiarism strategies and source texts to copy from. You should modify the source text according to my plagiarism strategies, and then output the modified text to me.Remember, to save your output time, you only need to output the modified text to me.Use <start> to indicate the start of the output, and <end> to indicate the end of the output.

\textbf{plagiarism strategy}: Utilize alternative words, modify the structure of sentences, rearrange the order of ideas, and alter the style of grammar. You can also adjust punctuation. This process might include dividing the original sentence into several distinct sentences, or conversely, merging several sentences into a single one. There are no limitations on the methods of modification.

\textbf{example}: Synonym replacement example:
Their bank account balance \textbf{reached the maximum insured amount} $\iff$ Their bank account balance was \textbf{at least 250 thousand dollars}

\textbf{Mr. Smith} just bought a new computer $\iff$ \textbf{Bob} just bought a new computer

The program \textbf{runs fast} $\iff$ The program \textbf{does not run slowly}

A spike in sales performance will \textbf{save the company from bankruptcy} $\iff$ Only a spike in sales performance \textbf{will halt the company’s bankruptcy}.

I \textbf{bought} a plane ticket online $\iff$ A plane ticket was \textbf{sold} to me online

Increase in \textbf{salaries} is often a great indicator of performance $\iff$ Increase in \textbf{salary} is often a great indicator of performance.

Is \textbf{this} your own work? $\iff$ Is \textbf{that} your own work?

There are many accounts of that hero’s legacy all \textbf{differing} in perspectives. $\iff$ There are \textbf{different} versions of that hero’s legacy.

\textbf{Grammatical structure modification example:}

Does working in that technology company pay well? Does it provide great 401k plans for its employees? $\iff$ They will work at the company to get high pay or to obtain great 401k plans.

How he would stare! $\iff$ He would surely stare!

Jacob \textbf{programmed} the game $\iff$ the game’s \textbf{programmer} is Jacob

“You must finish the project by the end of today,” \textbf{demanded my manager} $\iff$ \textbf{My manager demanded that} I must finish the project by the end of today

Alice presented \textbf{the gift to Bob}. $\iff$ Alice presented \textbf{Bob with the gift}.

All \textbf{spoken languages} are natural languages $\iff$ The \textbf{English language} is a natural language

\textbf{Initially,} we begin with the scientific method $\iff$ we begin with the scientific method \textbf{initially}

\textbf{Yesterday}, we were able to \textbf{complete our assignment and} submit it on time $\iff$ \textbf{Yesterday evening at 12:30 PM}, we were able to submit our assignment on time

\textbf{Semantic order modification Grammar style modification  example:}

I am building a dataset for plagiarism assignments and need your assistance. Please pretend to be a student who intends to plagiarize, and I will provide you with a plagiarism strategy and source text. Based on my plagiarism strategy, please modify the source text. Then output the modified text to me. Remember, to save your output time, you only need to output the modified text to me. $\iff$ Assistance is sought for the construction of a dataset tailored for plagiarism tasks. Imagine yourself in the shoes of a student poised to plagiarize. Your task is to apply this strategy, reshaping the given text accordingly. Once this transformation is complete, your sole responsibility is to present the revised text. The aim is to streamline the process; thus, only the altered text needs to be revealed. A strategy for such an act, along with the original text, will be furnished by me.

\textbf{source text}: Input your source text

\twocolumn
\newpage % 强制分页

\end{document}